\documentclass[11pt,a4paper,table]{article}
\usepackage[hyperref]{acl2020}
\usepackage{times}
\usepackage{latexsym}

\usepackage{microtype}
\usepackage{graphicx}
\usepackage{multirow}

\usepackage{listings}
\usepackage{xcolor}

\definecolor{codegreen}{rgb}{0,0.6,0}
\definecolor{codegray}{rgb}{0.5,0.5,0.5}
\definecolor{codepurple}{rgb}{0.58,0,0.82}
\definecolor{backcolour}{rgb}{0.95,0.95,0.92}

\lstdefinestyle{mystyle}{
    backgroundcolor=\color{backcolour},   
    commentstyle=\color{codegreen},
    keywordstyle=\color{magenta},
    numberstyle=\tiny\color{codegray},
    stringstyle=\color{codepurple},
    basicstyle=\ttfamily\footnotesize,
    breakatwhitespace=false,         
    breaklines=true,                 
    captionpos=b,                    
    keepspaces=true,                 
    numbers=left,                    
    numbersep=5pt,                  
    showspaces=false,                
    showstringspaces=false,
    showtabs=false,                  
    tabsize=2
}

\aclfinalcopy % Uncomment this line for the final submission
\title{Using multiple ASR hypotheses to boost i18n NLU performance}

\author{Charith Peris\hspace{0.5cm}Gokmen Oz\hspace{0.5cm}Khadige Abboud\hspace{0.5cm}Venkata sai Varada \AND Prashan Wanigasekara\hspace{0.5cm}Haidar Khan \\
\\
 Alexa AI, Amazon\\
 Cambridge MA \\
 \texttt{\{perisc, ogokmen, abboudk, vnk, wprasha, khhaida\}@amazon.com} \\
}

\date{}

\begin{document}
\maketitle
\begin{abstract}
Current voice assistants typically use the best hypothesis yielded by their Automatic Speech Recognition (ASR) module as input to their Natural Language Understanding (NLU) module, thereby losing helpful information that might be stored in lower-ranked ASR hypotheses. We explore the change in performance of NLU associated tasks when utilizing five-best ASR hypotheses when compared to status quo for two language datasets, German and Portuguese. To harvest information from the ASR five-best, we leverage extractive summarization and joint extractive-abstractive summarization models for Domain Classification (DC) experiments while using a sequence-to-sequence model with a pointer generator network for Intent Classification (IC) and Named Entity Recognition (NER) multi-task experiments. For the DC full test set, we observe significant improvements of up to 7.2\% and 15.5\% in micro-averaged F1 scores, for German and Portuguese, respectively. In cases where the best ASR hypothesis was not an exact match to the transcribed utterance (mismatched test set), we see improvements of up to 6.7\% and 8.8\% micro-averaged F1 scores, for German and Portuguese, respectively. For IC and NER multi-task experiments, when evaluating on the mismatched test set, we see improvements across all domains in German and in 17 out of 19 domains in Portuguese (improvements based on change in SeMER scores). Our results suggest that the use of multiple ASR hypotheses, as opposed to one, can lead to significant performance improvements in the DC task for these non-English datasets. In addition, it could lead to significant improvement in the performance of IC and NER tasks in cases where the ASR model makes mistakes. 
\end{abstract}

\section{Introduction}
\label{intro}

Recent years have seen a dramatic increase in the adoption of intelligent voice assistants such as Amazon Alexa, Apple Siri and Google Assistant. As use cases expand, these assistants are expected to process ever more complex user utterances and perform many different tasks. Some of the key components that enable the performance of these tasks are housed within the spoken language understanding (SLU) system; one being the Automatic Speech Recognition (ASR) module which transcribes the users' vocal sound wave into text and another being the Natural Language Understanding module which performs a variety of downstream tasks that help identify the actions requested by the user~\citep{ram2018conversational, gao2018neural}. These modules perform in tandem and are crucial for the successful processing of user utterances. Typical ASR models generate multiple hypotheses for an input audio signal, that are ranked by their confidence scores~\citep{li2020improving}. However, only the top ranked hypothesis (referred to hereafter as the ASR 1-best) is usually processed by the NLU module for downstream tasks~\citep{li2020improving}.
 
Three major tasks performed by the NLU module are Domain Classification (DC), Intent Classification (IC) and Named Entity Recognition (NER). DC predicts the domain relevant to the utterance (Weather, Shopping, Music etc.) and IC extracts actions requested by users (some examples are, buy an item, play a song or set a reminder). NER is focused on identifying and extracting entities from user requests (names, dates, locations, etc.). Current NLU models usually take in the ASR 1-best hypothesis as input to perform NLU recognition~\citep{li2020improving}. However, the highest-scored ASR hypothesis is not always correct and, at times, can lead to downstream failures including incorrect NLU hypotheses. These errors can be mitigated by utilizing multiple top-ranked ASR hypotheses (ASR n-best hypotheses) in NLU modeling, which have a higher likelihood of containing the correct hypothesis. Even in the case of all n-best hypotheses being incorrect, the NLU models may be capable of recovering the correct hypothesis by integrating the information contained within the n-best hypotheses. Hence, the use of multiple hypotheses should help obtain firmer predictions from ASR modules for their corresponding NLU module and result in improved performance. 
 
In this study we focus on two non-English internal datasets, German and Portuguese, and evaluate the use of ASR n-best hypotheses for improving NLU modeling within these contexts. Given that the ASR models we use in this experiment produce a maximum of five (or less) hypotheses per input utterance, we utilize all available hypotheses (referred to hereafter as the ASR 5-best) for our work. We leverage two BERT-based summarization models~\citep{devlin2018bert, liu2019a, liu2019b} and a sequence-to-sequence model with a pointer generator network~\citep{Rongali_2020} to extract the information from the ASR 5-best hypotheses. We show that using multiple hypotheses, as opposed to just one, can significantly improve the overall performance of DC, and the performance of IC and NER in cases where the ASR model makes mistakes. We describe relevant work in Section~\ref{sec:work} and present a description of our data set and opportunity cost analysis in Section~\ref{sec:data}. In Section~\ref{sec:exp} we describe the architecture of our models. In Section~\ref{sec:results}, we present our experimental results followed by our conclusions in Section~\ref{sec:conc}.

\section{Related work}
\label{sec:work}

Using deep learning models for summarization has been an active area of research in the recent past. Two popular types in current literature have been extractive summarization and abstractive summarization. Extractive summarization systems summarize by identifying and concatenating the most important sentences in a document whereas abstractive summarization systems conceptualize the task as a sequence-to-sequence problem and generate the summary by paraphrasing sections of the source document. Extensive work has been done on extractive summarization~\citep{liu2019a, cheng2016neural, nallapati2016summarunner, narayan-etal-2018-ranking, dong-etal-2018-banditsum, zhang-etal-2018-neural-latent, zhou-etal-2018-neural} and abstractive summarization~\citep{narayan2018dont, see2017point, alex2015neural, nallapati-etal-2016-abstractive} used in isolation. Furthermore, studies have shown improvement in summary quality when extractive and abstractive objectives have been used in combination~\citep{liu2019b, gehrmann-etal-2018-bottom, li-etal-2018-improving}. 

~\citet{liu2019a} proposed a simple, yet powerful, variant of BERT for extractive summarization in which they modified the input sequence of BERT from its original two sentences to multiple sentences. They used multiple classification tokens ([CLS]) combined with interval segment embeddings to distinguish multiple sentences within a document. They appended several summarization specific layers (either a simple classifier, a transformer or an LSTM) on top of the BERT outputs to capture document level features relevant for extracting summaries. Following this work,~\citet{liu2019b} proposed a model that comprises of the pre-trained BERT extractive summarization model~\citep{liu2019a} as the encoder and a decoder which consists of a 6-layered transformer~\citep{vaswani2017attention}. The encoder was fine-tuned in two stages, first on the extractive summarization task and then again on an abstractive summarization task resulting in a joint extractive-abstractive model that showed improved performance on summarization tasks.

The utilization of multiple ASR hypotheses for improved NLU model performance across DC, IC tasks was first introduced by~\citet{li2020improving}. They proposed the use of 5-best ASR hypotheses to train a BiLSTM language model, instead of using a single 1-best hypothesis selected using either majority vote, highest confidence score or a reranker. They explored two methods to integrate the n-best hypothesis: a basic concatenation of hypotheses text and a hypothesis embedding concatenation using max/avg pooling. The results show 14\%-25\% relative gains in both DC and IC accuracy. 

In our work, we explore the performance improvement offered by utilizing the ASR 5-best hypotheses in previously unexplored languages, German and Portuguese. We also differ from previous studies due to our use of the superior BERT-based extractive~\citep{liu2019a} and joint extractive-abstractive~\citep{liu2019b} summarization models to extract a summary hypothesis for the DC task, from the ASR 5-best.

Voice assistants traditionally handle IC and NER tasks using semantic parsing components which typically comprise of statistical slot-filling systems for simple queries and, in more recent time, shift-reduce parsers~\citep{gupta2018semantic, einolghozati2019improving} for more complex utterances.~\citet{Rongali_2020} proposed a unified architecture based on sequence-to-sequence models and pointer generator networks to handle both simple and complex IC and NER tasks with which they achieve state-of-the-art results. In this work, we use a model that expands this approach to consume the 5-best ASR hypotheses and evaluate its performance on IC/NER tasks for the two language datasets considered.

\section{Data}
\label{sec:data}

Our experiments focus on two non-English internal datasets; German and Portuguese. We run all utterances in each language through one language-specific ASR model and take the top-ranked ASR hypothesis for each utterance as ASR 1-best and all available hypotheses for each utterance (a maximum of five in our models) as ASR 5-best. In addition, we also obtain a human transcribed version of each utterance. For German, we use ~1.48 million utterances from 21 domains for training and validation. We split the data randomly within each domain, with 85\% used for training and 15\% for validation. An independent set of 193K utterances are used for testing. Within the independent test set we find 17K utterances where the ASR 1-best did not match the transcribed utterance exactly and mark them as the ``mismatched” test set. (Table~\ref{tab:tab1}). For Portuguese, we use 890K utterances from 19 domains for training and validation, split the same way as with German. Another 247K utterances are used for testing. We find 41K utterances within test, where the ASR 1-best did not match the transcribed utterance exactly, and mark them as the mismatched test set (Table~\ref{tab:tab1}).

\begin{table*}[]
\centering
\caption{Total data set sizes in terms of utterance counts}
\label{tab:tab1}
\resizebox{9cm}{!}{%
\begin{tabular}{|l|l|l|l|l|}
\hline
\rowcolor[HTML]{EFEFEF} 
Language & Train & Validation & Test (full) & Test (mismatched) \\ \hline
\cellcolor[HTML]{EFEFEF}German & 1,255,402 & 221,543 & 192,697 & 16,672 \\ \hline
\cellcolor[HTML]{EFEFEF}Portuguese & 756,148 & 133,438 & 246,638 & 40,896 \\ \hline
\end{tabular}%
}
\end{table*}

\subsection{Opportunity Cost Measurement}
\label{sec:opp_cost}

\citet{li2020improving} showed improvement in NLU model performance on English (en-US) upon utilizing the ASR 5-best hypotheses instead of only ASR 1-best. However, the impact of this on non-English languages has not yet been explored. To understand the opportunity of improvement that the ASR 5-best hypotheses can lend to NLU model performance in German and Portuguese datasets, we analyze the ASR 5-best hypotheses in comparison to the ground-truth human transcribed data for each of the considered language datasets. First, we calculate the number of exact matches to the transcribed utterance occurring in each of the top 5-best hypotheses. It should be mentioned that each ASR hypothesis is different from the others and only one hypothesis (if at all) can match the transcribed utterance. Next we compute the amount of exact matches found in the n$^{th}$-best hypothesis set, as a fraction of the volume of exact matches found at 1-best. The results are shown in Table~\ref{tab:tab2}. We find that the amount of exact matches that occur in 2-5 best hypotheses, compared to the volume of exact matches that occur in the top-ranked hypothesis, is large for Portuguese (30.16\%) and German (20.83\%) (see Table~\ref{tab:tab2}). This gives an indication of the opportunity present in using hypotheses beyond ASR 1-best for each language dataset. 

In Table~\ref{tab:tab3}, we further illustrate the use of the ASR 5-best hypotheses by showing three possible cases of stored information that we want our NLU model to extract; selecting the best matching hypothesis (first and second rows) and combining hypotheses (third row).

%\begin{figure}
%\centering
%\includegraphics[width=0.8\columnwidth]{fig1.png}
%\caption{Percentage of exact matches to the transcribed utterance occurring in the top 5 hypotheses for Portuguese and German}
%\label{fig:fig1}
%\end{figure}

%\begin{table}[]
%\centering
%\caption{Exact match and mean word error rates for the 5 best hypotheses in Portuguese and German}
%\label{tab:tab2}
%\resizebox{8cm}{!}{%
%\begin{tabular}{|
%>{\columncolor[HTML]{EFEFEF}}l |l|l|l|l|}
%\hline
%\cellcolor[HTML]{EFEFEF} & \multicolumn{2}{l|}{\cellcolor[HTML]{EFEFEF}Portuguese} & \multicolumn{2}{l|}{\cellcolor[HTML]{EFEFEF}German} \\ \cline{2-5} 
%\multirow{-2}{*}{\cellcolor[HTML]{EFEFEF} n-best} & \cellcolor[HTML]{EFEFEF}Exact Matches \% & \cellcolor[HTML]{EFEFEF}WER & \cellcolor[HTML]{EFEFEF}Exact Matches \% & \cellcolor[HTML]{EFEFEF}WER \\ \hline
%1 & 73.40\% & 1.53 & 80.70\% & 1.64 \\ \hline
%2 & 12.15\% & 1.43 & 8.28\% & 1.56 \\ \hline
%3 & 5.21\% & 1.86 & 4.04\% & 2 \\ \hline
%4 & 2.88\% & 2.01 & 2.69\% & 2.08 \\ \hline
%5 & 1.90\% & 2.13 & 1.80\% & 2.23 \\ \hline
%\end{tabular}%
%}
%\end{table}

\begin{table}[]
\centering
\caption{Exact Matches to the transcribed utterance found in ASR n-best as a percentage of Exact Matches found in ASR 1-best}
\label{tab:tab2}
\resizebox{5cm}{!}{%
\begin{tabular}{|
>{\columncolor[HTML]{EFEFEF}}l |l|l|}
\hline
\textbf{n} & \cellcolor[HTML]{EFEFEF}\textbf{Portuguese (\%)} & \cellcolor[HTML]{EFEFEF}\textbf{German (\%)} \\ \hline
\textbf{2} & 16.55 & 10.26 \\ \hline
\textbf{3} & 7.1 & 5.01 \\ \hline
\textbf{4} & 3.92 & 3.33 \\ \hline
\textbf{5} & 2.59 & 2.23 \\ \hline
\textbf{total} & 30.16 & 20.83 \\ \hline
\end{tabular}%
}
\end{table}

\begin{table*}[]
\centering
\caption{Illustrative examples in English that compares the 3-best ASR hypotheses to the transcribed utterance}
\label{tab:tab3}
\resizebox{10.5cm}{!}{%
\begin{tabular}{|l|l|l|l|}
\hline
\textbf{Transcription} & \textbf{1- best hypothesis} & \textbf{2-best hypothesis} & \textbf{3-best hypothesis} \\ \hline
\textbf{buy movie mystery} & \textbf{buy movie mystery}  &\textbf{buy} my tree & but move my tree \\ \hline
\textbf{who is nelson} & how is my son & \textbf{who is nelson }& how samsung \\ \hline
\textbf{play music} & pull \textbf{music} & pull news &\textbf{play} my muse \\ \hline
\end{tabular}%
}
\end{table*}

\section{Experimental Setup}
\label{sec:exp}

\subsection{DC models}
\label{sec:dc_mods}

For our DC experiments, we compare performance across the following classification models: 

\begin{itemize}
\item {\bf Baseline} – A BERT-based classification baseline model with MLP classifier trained on the {\it transcribed utterance} and tested on the {\it ASR 1-best}
\item  {\bf BSUMEXT}– A BERT-based extractive summarization model trained and tested on the {\it ASR 5-best}
\item  {\bf BSUMEXTABS}– A BERT-based joint extractive and abstractive summarization model trained and tested on the {\it ASR 5-best} 
\end{itemize}

Standard testing on transcribed utterances underestimates the combined ASR and NLU errors. In order to avoid this our test sets exclude transcribed utterances and thus reflect the real situation.

\begin{figure*}
\centering
\includegraphics[width=0.75\textwidth, trim={4cm 2cm 4cm 0cm}]{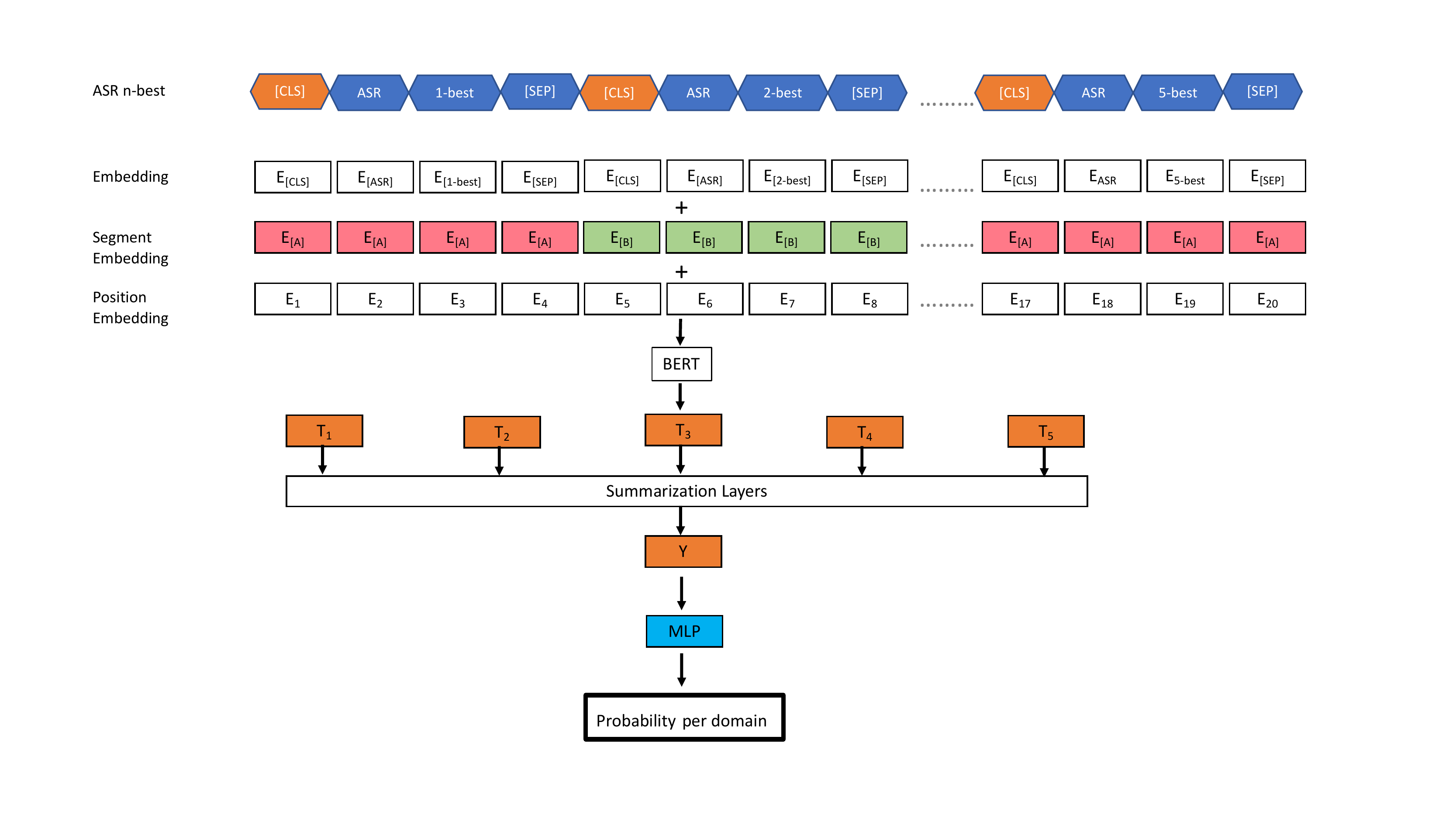}
\caption{A schematic of the architecture of the BSUMEXT}
\label{fig:fig2}
\end{figure*}

In Section~\ref{sec:data}, we described the simple extractive summarization model proposed by~\citet{liu2019a}. We adapt their extractive summarization model to take the ASR 5-best hypotheses as input and output a probability score per domain based on a summarized hypothesis. Figure~\ref{fig:fig2} shows the architecture of BSUMEXT with ASR 5-best input. The task of the BSUMEXT model is to create an extractive summary by picking from the class assigned to each hypothesis. This summary is then fed into a multi-layer perceptron classifier to perform the DC task. As in the case of~\citet{liu2019a}, vanilla BERT is modified to include multiple [CLS] symbols. Each symbol is used to obtained features of each of the ASR n-best hypotheses preceding it. Alternating hypotheses fed into the model are assigned a segment embedding (E\_A or E\_B), based on whether it is an even or odd numbered hypothesis. For example for a sentence ``play music'' :

\begin{lstlisting}
ASR 1-best: play muse     [E_A]
ASR 2-best: play mu chick [E_B]
ASR 3-best: play news     [E_A]
ASR 4-best: play mus      [E_B] 
ASR 5-best: play my sick  [E_A]
\end{lstlisting}

The model then takes the [CLS] representation of each ASR 5-best utterance and performs multi-headed attention to obtain the summary hypothesis.

For the BSUMEXTABS model, the BERT encoder is fine-tuned on an abstractive summarization task and then further fine-tuned on the extractive summarization task. In this model the summary hypothesis fed into the multi-layer perceptron classifier, is generated token by token in a sequence-to-sequence fashion. Similar to~\citet{liu2019b}, a decoupled fine-tuning schedule which separates the optimizers of the encoder and the decoder is used.

We trained each of our models for up to 30 epochs and use the best performing model, based on validation metrics, for evaluating the independent test set.

\subsection{IC/NER models}
\label{sec:icner_mods}

We compare the following models for the IC and NER tasks:

\begin{itemize}
\item {\bf Baseline} – A BERT-based classification baseline model trained on the {\it transcribed utterance} and tested on the {\it ASR 1-best}
\item {\bf BERT\_S2S\_NBEST\_PTR} – A BERT-based sequence-to-sequence model which employs a pointer generator network, trained on the {\it ASR 5-best + transcribed utterance} and tested on {\it ASR 5-best}

\end{itemize}

Instead of a typical sequence tagging problem,~\citet{Rongali_2020} propose a unified architecture to handle IC and NER tasks as a sequence generation problem. We build upon that approach. BERT\_S2S\_NBEST\_PTR is a sequence-to-sequence model augmented with a pointer generator network which functions as a self-attention mechanism. We expand the architecture proposed by~\citet{Rongali_2020} to include multiple input queries. The model task is to generate target words which can be either intent or slot delimiters or words that are from the source sequences. The pointer generator network enables the model to generate pointers to the source sequences (instead of using a large vocabulary of tokens) within the target sequence. An example of a source sequence with two ASR hypotheses and a target sequence looks as follows (we use spaces to delimit hypotheses and \_\&\_ to delimit separate tokens within an utterance):

\begin{lstlisting}
Source: ply_&_madonna play_&_mad_&_owner
Target: PlaySongIntent( @ptr1_0 ArtistName( @ptr0_1 )ArtistName )PlaySongIntent
\end{lstlisting}

where @ptr0\_1, for example, is a pointer to the second word ``madonna'' in the first utterance of the source query. One advantage of using pointers instead of the actual tokens is the smaller target vocabulary required for the decoder, resulting in a more light-weight model.

\begin{figure*}
\centering
\includegraphics[width=0.7\textwidth, trim={3cm 3cm 3cm 0cm}]{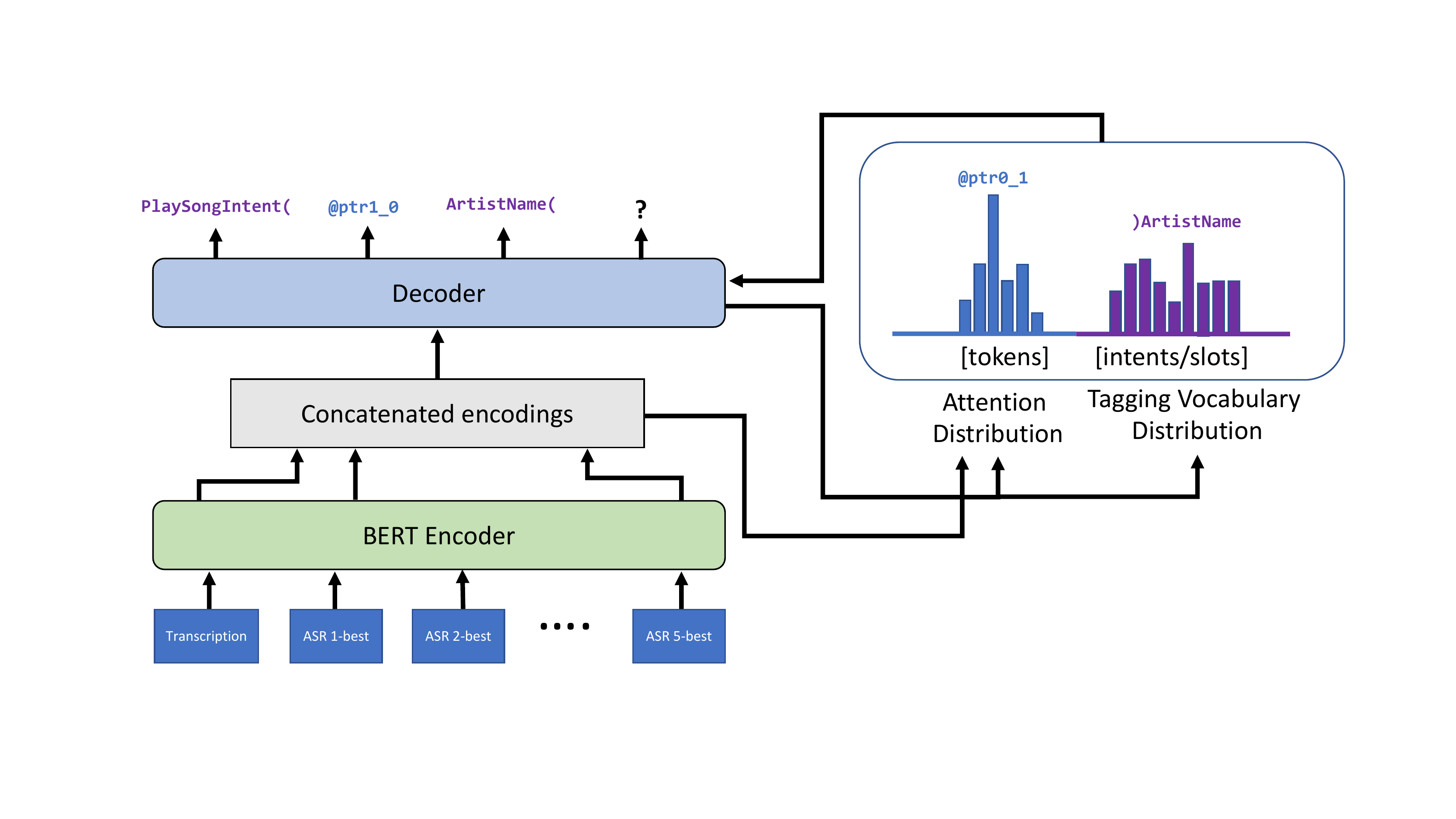}
\caption{A schematic of the sequence-to-sequence model with attention. Each ASR hypothesis is encoded separately. The encoder hidden states are then concatenated and passed to the decoder to have a cross-attention between encoder and decoder outputs over all ASR hypotheses.}
\label{fig:fig3}
\end{figure*}

The architecture consists of a pre-trained BERT encoder and a transformer decoder~\citep{devlin2018bert, vaswani2017attention}. The decoder is augmented with a pointer generator network that functions as a self-attention mechanism. Figure~\ref{fig:fig3} shows the high-level architecture. The Bert encoder processes each ASR hypothesis separately. The encoder hidden states over all ASR hypotheses are then concatenated and passed to the decoder. The decoder hidden states are used to update the attention mechanism and the tagging vocabulary and pointer distributions (see~\citet{Rongali_2020} for detailed descriptions). These probability distributions of tags and pointers are used to determine the next word and tag that is output by the decoder. The model is trained by minimizing sequence cross entropy loss over the training set.

These models are domain-specific multi-task models which handle both IC and NER tasks simultaneously. We trained one model per domain with all models trained for up to 50 epochs. The best performing model based on validation metrics was used for evaluating the independent test set.

\section{Results and Discussion}
\label{sec:results}

\subsection{Evaluation}
\label{sec:eval}

We measure the success of our DC experiments by comparing both micro- and macro-averaged F1 scores of our experimental models to those of the baseline model. Micro- and macro-averaged F1 scores are defined as 

\begin{equation}
\label{eq:eq0a}
\resizebox{0.3\hsize}{!}{$F1_{micro} = \frac{2 \times P \times R }{P + R}$}
\end{equation}

\begin{equation}
\label{eq:eq0b}
\resizebox{0.6\hsize}{!}{$F1_{macro} = \frac{1}{n} \sum_{i} F1_{i} = \frac{1}{n} \sum_{i} \frac{2\times P_{i}\times R_{i}} {P_{i} + R_{i}}$}
\end{equation}
where $P$ and $R$ are overall precision and recall respectively and $P_{i}$ and $R_{i}$ are the within class precisions and recalls respectively. We also calculate the relative change in error of each experimental model run with respect to baseline as shown in equation~\ref{eq:eq1}. Note that ``lower-is-better'' for this metric. In addition to these metrics calculated on the {\it full} test data set, we also calculate these metrics on the {\it mismatched} test set utterances where the ASR 1-best did not match the transcribed utterance.

\begin{equation}
\label{eq:eq1}
\resizebox{1\hsize}{!}{$\Delta_{err} = 100 \times \frac {( (100 - F1_{experiment}) - (100 - F1_{baseline}) )} {(100 - F1_{baseline}) )}$}
\end{equation}

For the IC and NER experiments, we use Semantic Error Rate (SemER)~\citep{su2018reranker} as our metric of choice. SemER is defined as follows:

\begin{equation}
\label{eq:eq2}
\resizebox{0.4\hsize}{!}{$SemER = \frac{D + I + S}{C+ D + S}$}
\end{equation}

where D=deletion, I=insertion, S=substitution and C=correct-slots. The Intent is treated as a slot in this metric and Intent error, considered as a substitution. We use the relative change in SemER with respect to the baseline model (equation~\ref{eq:eq3}), both overall and per domain in order to evaluate the success of our models. Note that “lower-is-better” for relative change in SemER as well. 

\begin{equation}
\label{eq:eq3}
\resizebox{1\hsize}{!}{$\Delta_{sem} = 100 \times \frac{( SemER_{experiment} - SemER_{baseline} )}{ SemER_{baseline} }$}
\end{equation}

\subsection{DC experiments}
\label{sec:dc_exp}

\begin{table}[]
\centering
\caption{Evaluation on the {\it full} and {\it mismatched} test sets for DC. Relative change in error rate ($\Delta_{err}$) measured against baseline for each metric is shown in each succeeding column (negative is good).  }
\label{tab:dc_results}
\resizebox{0.5\textwidth}{!}{%
\begin{tabular}{|
>{\columncolor[HTML]{EFEFEF}}l |l|l|l|l|}
\hline
\cellcolor[HTML]{EFEFEF} & \multicolumn{2}{l|}{\cellcolor[HTML]{EFEFEF}\textbf{Full set}} & \multicolumn{2}{l|}{\cellcolor[HTML]{EFEFEF}\textbf{Mismatched set}} \\ \cline{2-5} 
\multirow{-2}{*}{\cellcolor[HTML]{EFEFEF}\textbf{Model}} & \cellcolor[HTML]{EFEFEF}\textbf{\begin{tabular}[c]{@{}l@{}}f1\_micro \\ ($\Delta_{err}$)\end{tabular}} & \cellcolor[HTML]{EFEFEF}\textbf{\begin{tabular}[c]{@{}l@{}}f1\_macro \\  ($\Delta_{err}$)\end{tabular}} & \cellcolor[HTML]{EFEFEF}\textbf{\begin{tabular}[c]{@{}l@{}}f1\_micro \\ ($\Delta_{err}$)\end{tabular}} & \cellcolor[HTML]{EFEFEF}\textbf{\begin{tabular}[c]{@{}l@{}}f1\_macro \\  ($\Delta_{err}$)\end{tabular}} \\ \hline
\multicolumn{5}{|c|}{\cellcolor[HTML]{EFEFEF}\textbf{German}} \\ \hline
\textbf{BSUMEXT} & -1.60\% & -4\% & -5.40\% & -12\% \\ \hline
\textbf{BSUMEXTABS} & -7.20\% & -3.90\% & -6.70\% & -2.30\% \\ \hline
\multicolumn{5}{|c|}{\cellcolor[HTML]{EFEFEF}\textbf{Portuguese}} \\ \hline
\textbf{BSUMEXT} & -12.60\% & 4.90\% & -6.30\% & -0.30\% \\ \hline
\textbf{BSUMEXTABS} & -15.50\% & -7.30\% & -8.80\% & -7.40\% \\ \hline
\end{tabular}%
}
\end{table}

Table~\ref{tab:dc_results} describes the performance of all the models defined in Section 4.1 on the full test set and the mismatched test set (see Section~\ref{sec:data} and Table~\ref{tab:tab1}). The full test set enables us to understand the general performance improvement that can be achieved by using summarization models. Although utilizing the full ASR 5-best hypotheses might offer some improvement even in cases where the ASR 1-best hypothesis is an exact match to the transcribed utterance, much more value-add is expected when using the ASR 5-best hypotheses in cases where there is a mismatch between the transcribed utterance and ASR 1-best. To study this use case, we use the mismatched test set. 
 
We observed that a majority of F1 scores across all models for German exceeded their corresponding values in Portuguese. Our opportunity cost analysis showed that exact matches between the transcribed utterance and ASR 2-5-best for Portuguese are higher than for German (see Section~\ref{sec:opp_cost}). This suggests that the German ASR model tends to perform better than the Portuguese ASR model. In this light, the smaller gains in relative change in error observed for German when compared to Portuguese are likely due to the German ASR model being superior and therefore leaving smaller room for improvement.

\begin{figure*}
\centering
\includegraphics[width=0.9\textwidth]{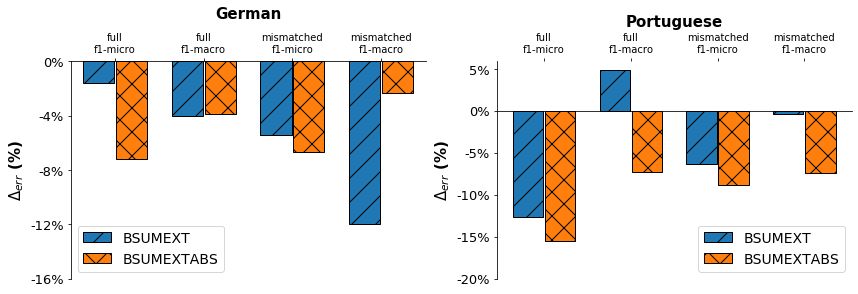}
\caption{Relative change in error rate measured against baseline for each metric on full and unmatched test sets for DC experiments.}
\label{fig:fig4}
\end{figure*}

Figure~\ref{fig:fig4} displays the relative changes of each model against the baseline for each dataset. When considering micro-averaged F1 scores, the BSUMEXT and BSUMEXTABS models out-perform the baseline in all cases, with the later out-performing the former. This shows that the use of ASR 5-best hypotheses can significantly improve overall classification for both language datasets. The BSUMEXTABS models also consistently out-perform the baseline on macro-averaged F1 scores, showing improvement in mean within-class classification scores as well. This suggests that BSUMEXTABS with additional fine-tuning on the abstractive task, is in general more successful at creating a firmer hypothesis for DC than the pure extractive summarization of BSUMEXT. For Portuguese, even with the relatively large percentage of exact matches available for extraction within its ASR 2-5 hypotheses (see Section 3), BSUMEXTABS consistently outperforms BSUMEXT across all metrics and datasets.

\begin{table}[]
\centering
\caption{Joint evaluation on {\it full} and {\it mismatched} test sets for IC/NER tasks. $\Delta_{sem}$ (\%) is the relative change in SemER against baseline for each domain (negative is good).}
\label{tab:icner-results}
\resizebox{0.45\textwidth}{!}{%
\begin{tabular}{|
>{\columncolor[HTML]{EFEFEF}}l |l|l|}
\hline
\multicolumn{3}{|l|}{\cellcolor[HTML]{EFEFEF}\textbf{German}} \\ \hline
\cellcolor[HTML]{EFEFEF} & \multicolumn{2}{l|}{\cellcolor[HTML]{EFEFEF}\textbf{S2S\_NBEST\_PTR}} \\ \cline{2-3} 
\multirow{-2}{*}{\cellcolor[HTML]{EFEFEF}\textbf{Domain}} & \cellcolor[HTML]{EFEFEF}\textbf{Full Set $\Delta_{sem}$ (\%)} & \cellcolor[HTML]{EFEFEF}\textbf{Mismatched   Set $\Delta_{sem}$ (\%)} \\ \hline
domain A & 14.79 & -14.16 \\ \hline
domain B & 30.33 & -10.27 \\ \hline
domain C & 25.25 & -5.83 \\ \hline
domain D & 95.41 & -7.3 \\ \hline
domain E & 16.38 & -12.68 \\ \hline
domain F & 12.54 & -18.9 \\ \hline
domain G & -33.51 & -23.2 \\ \hline
domain H & 7.41 & -14.2 \\ \hline
domain I & 12.96 & -25.2 \\ \hline
domain J & 15.27 & -3.72 \\ \hline
domain K & 32.02 & -7.42 \\ \hline
domain L & 89.45 & -18.63 \\ \hline
domain M & 643.85 & -15.95 \\ \hline
domain N & 1.06 & -7.21 \\ \hline
domain O & -34.8 & -25.02 \\ \hline
domain P & 26.52 & -8.74 \\ \hline
domain Q & 8.47 & -6.13 \\ \hline
domain R & 69.35 & -13.76 \\ \hline
domain S & 19.07 & -2.12 \\ \hline
domain T & -4.25 & -10.93 \\ \hline
domain U & 1.92 & -7.33 \\ \hline
\textbf{Overall} & 19.17 & -11.64 \\ \hline
\multicolumn{3}{|l|}{\cellcolor[HTML]{EFEFEF}\textbf{Portuguese}} \\ \hline
\cellcolor[HTML]{EFEFEF} & \multicolumn{2}{l|}{\cellcolor[HTML]{EFEFEF}\textbf{S2S\_NBEST\_PTR}} \\ \cline{2-3} 
\multirow{-2}{*}{\cellcolor[HTML]{EFEFEF}\textbf{Domain}} & \cellcolor[HTML]{EFEFEF}\textbf{Full Set $\Delta_{sem}$ (\%)} & \cellcolor[HTML]{EFEFEF}\textbf{Mismatched   Set $\Delta_{sem}$ (\%)} \\ \hline
domain A & 2.89 & -14.11 \\ \hline
domain B & 18.88 & -7.94 \\ \hline
domain C & 46.86 & -14.7 \\ \hline
domain D & 4.3 & 3.16 \\ \hline
domain E & -12.54 & -30.65 \\ \hline
domain F & 5.87 & -18.89 \\ \hline
domain G & 6.56 & -3.7 \\ \hline
domain H & 24.64 & -2.57 \\ \hline
domain I & 71.12 & -24.42 \\ \hline
domain J & -7.69 & -10.03 \\ \hline
domain K & 19.16 & -5.45 \\ \hline
domain L & 11.15 & -9.97 \\ \hline
domain M & 3.54 & -10.58 \\ \hline
domain N & 48.85 & -10.15 \\ \hline
domain O & -30.38 & -59.98 \\ \hline
domain P & 6.84 & -12.29 \\ \hline
domain Q & 0.11 & -15.66 \\ \hline
domain R & 20.49 & -8.94 \\ \hline
domain V & 1533.33 & 47.62 \\ \hline
\textbf{Overall} & 106.58 & -8.09 \\ \hline
\end{tabular}%
}
\end{table}

\subsection{IC and NER experiments}
\label{sec:icner_exp}

Table~\ref{tab:icner-results} describes the performance of all the models defined in Section~\ref{sec:icner_mods} on domain-level data from the full test set and the mismatched test set. As with the DC experiments, we use the full test set to understand the general overall performance improvement, and use the mismatched test set to identify improvement in cases where the ASR 1-best hypothesis is not an exact match to the transcribed utterance.

When evaluating the BERT\_S2S\_NBEST\_PTR model, we find that it tends improve performance specifically on the mismatched test set. For German, we find improved performance across every domain on the mismatched test set (see Table~\ref{tab:icner-results}) with an overall SemER improvement of 11.6\% against baseline. However, we only observe improvement in three domains on the full set, while other domains show degradation in SemER. It is also interesting to note that the domains that improve also had low utterance counts. For Portuguese, testing on the mismatched test set yields improved performance across 17 out of 19 domains (see Table~\ref{tab:icner-results}) with an overall SemER improvement of 8.1\% against baseline, while we see only three domains show improvement on the full test set. Our results suggest that the ASR 1-best hypothesis works well for IC/NER tasks. The noise added by additional hypotheses seem to degrade results in the general use case. However, the additional hypotheses tend to be very helpful in cases where the ASR model makes mistakes (i.e. mismatched set data where the ASR 1-best is not an exact match to the transcribed utterance). 

Our full test set results show that the baseline model appears to be a better choice for the IC/NER tasks. However, if we could detect user utterances where the ASR model might have made a mistake in its top hypothesis, the ASR outputs (i.e. the set of all hypotheses) of these utterances could be channeled to a separate NLU model such as BERT\_S2S\_NBEST\_PTR, that could build a better hypothesis than the baseline and improve overall IC/NER performance. 

\begin{figure*}
\centering
\includegraphics[width=0.9\textwidth]{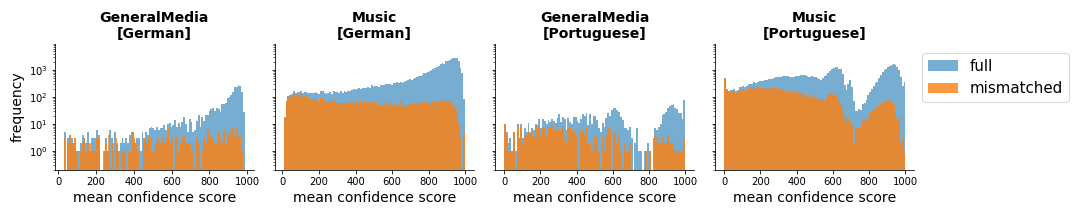}
\caption{Frequency distributions of mean confidence score across all available hypotheses for each data point in {\it full} and {\it mismatched} test sets. We show results for only two domains for each language due to space limitations. The distributions show similar shape across all domains within each language dataset. }
\label{fig:fig5}
\end{figure*}

We analyzed the confidence scores of our ASR models on the full and mismatched test set hypotheses to explore the possibility of detecting a mismatched set ASR output. For each ASR output we obtain the mean confidence score across all available hypotheses. We then compare the frequency distributions of the mean confidence scores in the full and mismatched test sets. Figure~\ref{fig:fig5} shows the resulting distributions for two example domains for each language dataset. We find that the full set shows a strong peak at high confidence scores while the mismatched set shows a more uniform distribution. The pronounced difference in distribution shape suggests that a thresholding mechanism based on the confidence score output by the ASR model (or a simple classifier trained on ASR outputs and scores) might be used to predict mismatched test set outputs with good confidence. Leveraging such a mechanism might enable the use of a second model such as BERT\_S2S\_NBEST\_PTR to improve performance in these mismatched cases, and in turn improve overall IC/NER performance.

\section{Conclusions and future work}
\label{sec:conc}

In this study, we explore the benefits of using ASR 5-best hypotheses for the NLU tasks in the German and Portuguese datasets. We explore several models to perform DC and IC/NER tasks and evaluate their performance against baseline models that use ASR 1-best. We find significant overall improvement in performance for the DC task. We also find significant improvement in performance of the jointly evaluated IC/NER tasks in cases where the ASR 1-best hypothesis is not an exact match to the transcribed utterance. For the DC task, our results suggest that the use of ASR 5-best helps produce better hypotheses and thereby greater improvements in the case of slight lower quality ASR models.

Our next steps will include exploring how different data splits based on ASR confidence scores might affect the sequence-to-sequence model performance. Furthermore, we will explore performance improvements in IC and NER tasks, using different model architectures and training schedules. We will also expand our study to a larger set of languages in order to understand how the use of multiple ASR hypotheses might affect languages with different lexical distributions. Languages which use multiple scripts (Japanese, Hindi, Arabic etc.) or which are more opaque and likely to have heterographs (e.g., “serial, “cereal”) and those that have less standardized spelling systems (Hindi etc) are more likely to have ASR errors. They may have different levels of improvement with the use of ASR 5-best hypotheses and we hope to analyze this in our future work.

\section*{Acknowledgments}
We thank Saleh Soltan for extensive discussions on the BERTSUM model architecture and implementation as well as for creating the BERT embeddings and making them available for our use. We thank Xinyue Liu for valuable contributions to discussions on baselines and model optimization. We thank Mukund Harakere Sridhar for help with pretraining the original encoders, extensions of which were used in this work. We also thank Karolina Owczarzak, Chengwei Su and Wael Hamza for helpful discussions and advice on this work.

\bibliography{nbest}
\bibliographystyle{acl_natbib}

\appendix

\end{document}